\documentclass[conference]{IEEEtran}
\IEEEoverridecommandlockouts
\usepackage{cite}
\usepackage{booktabs}
\usepackage{amsmath,amssymb,amsfonts}
\usepackage{algorithmic}
\usepackage{graphicx}
\usepackage{textcomp}
\usepackage{xcolor}
\usepackage{hyperref}
\def\BibTeX{{\rm B\kern-.05em{\sc i\kern-.025em b}\kern-.08em
    T\kern-.1667em\lower.7ex\hbox{E}\kern-.125emX}}

\usepackage{tikz}

\newcommand\submittedtext{%
  \footnotesize This work has been accepted to the 2024 International Conference on Machine Learning and Applications (ICMLA). This work has been submitted to the IEEE for possible publication. Copyright may be transferred without notice, after which this version may no longer be accessible.}

\newcommand\submittednotice{%
\begin{tikzpicture}[remember picture,overlay]
\node[anchor=south,yshift=10pt] at (current page.south) {\fbox{\parbox{\dimexpr0.65\textwidth-\fboxsep-\fboxrule\relax}{\submittedtext}}};
\end{tikzpicture}%
}
\renewcommand\fbox{\fcolorbox{red}{white}}
\begin{document}

\title{Developing the Temporal Graph Convolutional Neural Network Model to Predict Hip Replacement using Electronic Health Records\\
}

\author{
    \IEEEauthorblockN{Zoe Hancox}
    \IEEEauthorblockA{
        \textit{School of Computer Science} \\
        \textit{University of Leeds}\\
        Leeds, United Kingdom \\
        \href{mailto:z.l.hancox@leeds.ac.uk}{z.l.hancox@leeds.ac.uk}
    }
    \vspace{-1em}
    \and
    \IEEEauthorblockN{Sarah R. Kingsbury}
    \IEEEauthorblockA{
        \textit{Leeds Institute of Rheumatic and Musculoskeletal Medicine (LIRMM)} \\ 
        \textit{NIHR Leeds Biomedical Research Centre (BRC)} \\
        \textit{University of Leeds}\\
        Leeds, United Kingdom \\
        \href{mailto:s.r.kingsbury@leeds.ac.uk}{s.r.kingsbury@leeds.ac.uk}
    }
    \vspace{-1em}
    \and
    \IEEEauthorblockN{Andrew Clegg}
    \IEEEauthorblockA{
        \textit{Academic Unit for Ageing and Stroke Research} \\
        \textit{Bradford Institute for Health Research}\\
        Bradford, United Kingdom \\
        \href{mailto:a.p.clegg@leeds.ac.uk}{a.p.clegg@leeds.ac.uk}
    }
    \vspace{-5em}
    \and
    \IEEEauthorblockN{Philip G. Conaghan}
    \IEEEauthorblockA{
        \textit{LIRMM} \\ 
        \textit{NIHR Leeds BRC} \\
        \textit{University of Leeds}\\
        Leeds, United Kingdom \\
        \href{mailto:p.conaghan@leeds.ac.uk}{p.conaghan@leeds.ac.uk}
    }
    \vspace{-5em}
    \and
    \IEEEauthorblockN{Samuel D. Relton}
    \IEEEauthorblockA{
        \textit{Division of Health Services Research} \\
        \textit{University of Leeds}\\
        Leeds, United Kingdom \\
        \href{mailto:s.d.relton@leeds.ac.uk}{s.d.relton@leeds.ac.uk}
    }
}


\maketitle
\submittednotice

\begin{abstract}
    \textbf{Background:} Hip replacement procedures improve patient lives by relieving pain and restoring mobility. Predicting hip replacement in advance could reduce pain by enabling timely interventions, prioritising individuals for surgery or rehabilitation, and utilising physiotherapy to potentially delay the need for joint replacement. This study predicts hip replacement a year in advance to enhance quality of life and health service efficiency.
    \textbf{Methods:} Adapting previous work using Temporal Graph Convolutional Neural Network (TG-CNN) models, we construct temporal graphs from primary care medical event codes, sourced from ResearchOne EHRs of 40-75-year-old patients, to predict hip replacement risk. We match hip replacement cases to controls by age, sex, and Index of Multiple Deprivation. The model, trained on 9,187 cases and 9,187 controls, predicts hip replacement one year in advance. We validate the model on two unseen datasets, recalibrating for class imbalance. Additionally, we conduct an ablation study and compare against four baseline models.
    \textbf{Results:} Our best model predicts hip replacement risk one year in advance with an AUROC of 0.724 (95\% CI: 0.715-0.733) and an AUPRC of 0.185 (95\% CI: 0.160-0.209), achieving a calibration slope of 1.107 (95\% CI: 1.074-1.139) after recalibration.
    \textbf{Conclusions:} The TG-CNN model effectively predicts hip replacement risk by identifying patterns in patient trajectories, potentially improving understanding and management of hip-related conditions.
\end{abstract}


\section{Introduction}

The ageing population, coupled with increasing obesity rates, contributes to a rising prevalence of Osteoarthitis (OA) and hip replacements, causing pain and diminished quality of life \cite{WhittyCMOReport2020,chen2023temporal,Conaghan2014,Arslan2022KneeOAIncidence,Cook2023FrailtyDeprivation}. OA presents a significant challenge for UK healthcare, with musculoskeletal (MSK) issues accounting for 1 in 5 to 1 in 7 GP consultations \cite{Keavy2022MSKPrev,VersusArthritis2023}. The burden is exacerbated by a growing number of joint replacements, particularly for knee and hip OA \cite{NJR2015}, placing strain on both the National Health Service (NHS) and affected individuals \cite{Hobbs2016ClinicalWorkload}. For patients with advanced OA, total joint replacement may be recommended if conservative treatments fail.

MSK conditions impose not only financial burdens but also significant well-being impacts, potentially leading to conditions like depression or obesity \cite{Keavy2022MSKPrev}. Previous analysis of CPRD data revealed that 32.5\% of individuals with hip OA underwent total joint replacement \cite{Cook2023FrailtyDeprivation}. Between 2003 to 2014, there were 708,311 primary total hip replacements in the UK \cite{NJR2015}. Recognising alternative diagnoses associated with hip replacement surgeries may help delay or reduce the need for these invasive procedures.

Electronic Health Record (EHR) data, commonly collected during primary and secondary healthcare visits, provides structured and temporal information. However, the irregular time intervals within EHRs present a challenge for designing predictive algorithms in healthcare. Predicting post-surgery intensive care unit admission or hip replacement in advance using preoperative data supports efficient resource allocation and improves patient outcomes. Using temporal EHRs could assist in accurately forecasting health-related outcomes \cite{Zhang2019ATTAIN, Wu2022OA-MedSQL}, thereby aiding clinical decision-making and patient care. Additionally, machine learning models extract valuable insights, such as disease progression, from EHRs \cite{Chiew2020}.

This paper presents a model combining temporal graphs, 3D convolutional neural networks (CNNs), and Long Short-term Memorys (LSTMs) to process Read Codes and time intervals from EHRs. Read Codes are clinical terms used in the UK to electronically record patient symptoms, results and treatments in health and social care. The 3D CNN captures short-term temporal patterns, while the LSTM handles longer-term associations. The project aims to explore variations of the \texttt{Temporal Graph Convolutional Neural Network} (\texttt{TG-CNN}) model for predicting hip replacements in individuals \cite{Hancox2022}. The long-term goal is to develop a clinical decision support tool promoting trust, aiding patient-clinician discussions, offering personalised medicine, and enhancing patient safety.

This paper makes the following contributions:

\begin{itemize}
    \item Representing individual patient EHRs as temporal graphs and adapting the TG-CNN model to the clinical domain.
    \item Improving the efficiency of the 3D CNN layer using sparse linear algebra.
    \item Conducting an ablation study to evaluate model layers, regularisation techniques, and inclusion of demographic and prescription data.
    \item Developing a prognostic model for predicting hip replacement risk one year in advance.
\end{itemize}

\section{Related Work}

\cite{Yu2019HipKneeReplace} investigated predicting the risk of hip 
replacement among patients with osteoarthritis or hip
pain in the UK using CPRD data. They developed cumulative incidence equations to forecast the 10-year probability of hip replacement, narrowing down the model to 20 predictors for hip replacement. They achieved an Area under the Receiver Operator Curve (AUROC) of 0.72 and a C-slope of 1.00.
This study marked the first clinical risk prediction model for hip 
replacement in patients newly presenting with hip pain/osteoarthritis in primary care. To the best of our knowledge this is the only existing work to predict hip replacement risk without using radiographic images.

Hip injury strongly indicated future hip replacement risk \cite{Yu2019HipKneeReplace}. Other models, such as those by \cite{Leung2020} and \cite{Yi2024}, employ radiographic images for prediction. Utilising \texttt{TG-CNN} and EHR data from primary care services to predict hip replacement risk could reveal novel patterns associated with the procedure, thus enhancing predictive accuracy. This approach might facilitate early intervention or triage for individuals showing potential hip replacement indicators.

In a study on deep learning methods for handling irregularly sampled medical time series data \cite{Sun2020IrregularTimeReview}, linear regression, Random Forests (RFs), and Support Vector Machines (SVMs) showed predictive capabilities for health outcomes but struggled to integrate irregular time data effectively. Conversely, T-LSTM and DATA-Gated Recurrent Unit (GRU) models were suitable for accommodating irregularly sampled data, with DATA-GRU featuring implicit attention mechanisms that enhance interpretability. Our model generates informative features, evaluates events within their clinical context to reduce reporting bias, and incorporates elapsed time between events.

The \texttt{TG-CNN} model, originally demonstrated on an online course clickstream dataset to predict student dropout \cite{Hancox2022}, faces a key distinction in the medical domain when using EHRs: EHRs can record multiple Read Codes per visit, leading to parallelism within graph networks, whilst click actions can only occur sequentially.

\section{Methods}

The methodology proposed in this paper incorporates the elapsed time between EHR events (CTV3 Read Codes), alongside automatically capturing informative features within the data. In this work, we build on the \texttt{TG-CNN} model which combines temporal graph theory and CNNs to analyse structural, temporal data \cite{Hancox2022}.

\subsection{Cohort Analysis} 

We utilise NHS data from ResearchOne, managed by The Phoenix Partnership (TPP), comprising clinical and administrative data from 151,565 patients aged 40-75, attending healthcare practices in England that use SystmOne. Patients had their first record of joint pain clinically coded between April 1st, 1999, and March 31st, 2014 \cite{Smith2020UnderstandingEngland}. This age range was chosen as these patients are likely to present with MSK symptoms. Descriptive cohort characteristics are detailed in Table \ref{tab:stats of overall EHR dataset}, prior to removing ineligible records.

\begin{table}
    \centering
    \caption{Statistical summary of the EHR dataset.}
    \begin{tabular}{lr}
        \toprule
        Data collection period & 01/04/1999 - 31/03/2014\\ 
        Patient age on 01/04/1999 (mean (SD)) & 56.67 (9.37)\\
        Total \# of patients & 151,565\\
        Total \# of visits & 243,700\\
        Average \# of visits per patient & 1.827\\
        \midrule
        Total \# of unique EHR codes & 2,353\\
        Average \# of EHR codes per visit & 1.36\\
        Max \# of EHR codes per visit & 10\\
        Max \# of EHR codes recorded for a patient & 534 \\
        Average \# of visits across the patients & 19.99 \\
        Max \# of visits for one patient & 423 \\
        \# of patients who have a hip replacement & 12,978 \\
    \bottomrule
    \end{tabular}
    \label{tab:stats of overall EHR dataset}
\end{table}

In this paper, an EHR refers to a patient's record, consisting of reported primary care CTV3 Read Codes with timestamps. Each record also includes age, sex, and Index of Multiple Deprivation (IMD) score, a measure of location-based deprivation based on postcodes \cite{IMD2019}. Including IMD in our model is crucial, as research indicates that individuals with higher IMD scores (indicating lower deprivation) are less likely to have hip replacements \cite{Cook2023FrailtyDeprivation}.

\subsection{Data Extraction}

\begin{figure}[h]
    \centering
    \includegraphics[width=1\linewidth]{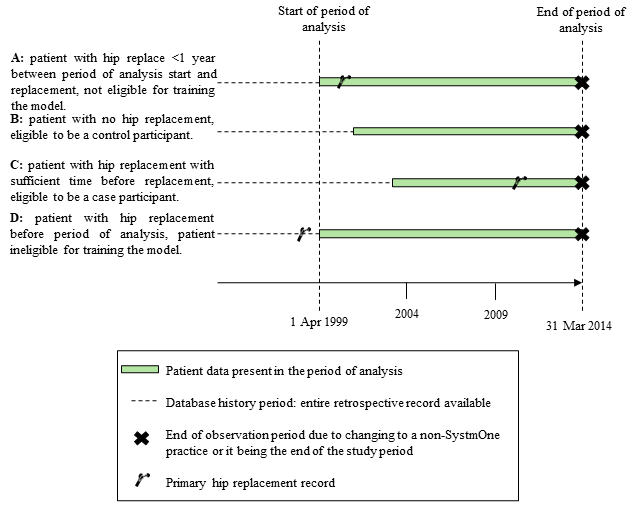}
    \caption{
    Four patient examples (A-D) and their eligibility for the study cohort. 
    Historical records for these patients were included where available. Patient data were included from the start of the analysis period or from their entry into the database if it occurred after April 1, 1999 (B and C). Patients were followed until the end of the analysis period or until they changed to a practice not using SystmOne. A primary hip replacement event was considered incident if the first hip replacement was recorded within the analysis period (C).
}
    \label{fig:one_year_windowing}
\end{figure}

\textbf{Time windowing:} 
To forecast hip replacements occurring one year in advance, we identify the specific time of the replacement and gather all available patient records from 1999 up to one year prior to the replacement date (see Figure \ref{fig:one_year_windowing}).

\textbf{Patient inclusion criteria:} Patients needed a minimum of two primary care provider visits before the one year prior to a hip replacement (partial or full). Clinician-selected CTV3 Read Codes (n=45) for primary hip replacement were utilised for the outcome, with the incident date defined as the first occurrence of one of these codes. The codes can be found in supplementary materials\footnote{\url{https://github.com/ZoeHancox/Sparse_TGCNN}} Table S1. The initial recorded hip replacement instance served as the label.
To mitigate the influence of age, sex, and IMD on joint replacement, one-to-one matching was performed for these variables in the training dataset of each hip replacement patient. This ensures prediction performance is unaffected by differences in these variables.
A small fraction (0.007\%, n=1,039) of patients in this dataset deceased, with 7 of them having undergone hip replacement. 

\subsection{Feature Choices} 
\label{sub:feature_choices}

Demographic characteristics, including sex, age, and IMD, were collected alongside all events and medical history within specified time periods. The most commonly used 512 Read Codes, covering 99.71\% of the data, were selected to construct temporal graph-based EHR representations for each patient, ensuring detailed representation while avoiding overfitting. Additionally, body mass index (BMI) was included as a predictor among the Read Codes. 
The first recorded IMD value for each patient was used for consistency, given its stability over time. Predictors synonymous with hip replacement, such as Read Codes for hospital referral, were excluded from the model. 
For the ablation study, medications were appended as 6 predictors using British National Formulary (BNF) codes, grouping drugs into opioids (04.07.02), non-opioid analgesics (04.07.01), and non-steroid anti-inflammatory drug (NSAIDs) (10.01.01), further subgrouped into acute and repeat prescription types. This addition brings the total number of predictors in the model to 521 when we include age, sex and IMD score.

\subsection{Temporal Graph Representation of EHRs} 

\begin{figure*}[h]
    \centering
    \includegraphics[width=0.9\linewidth]{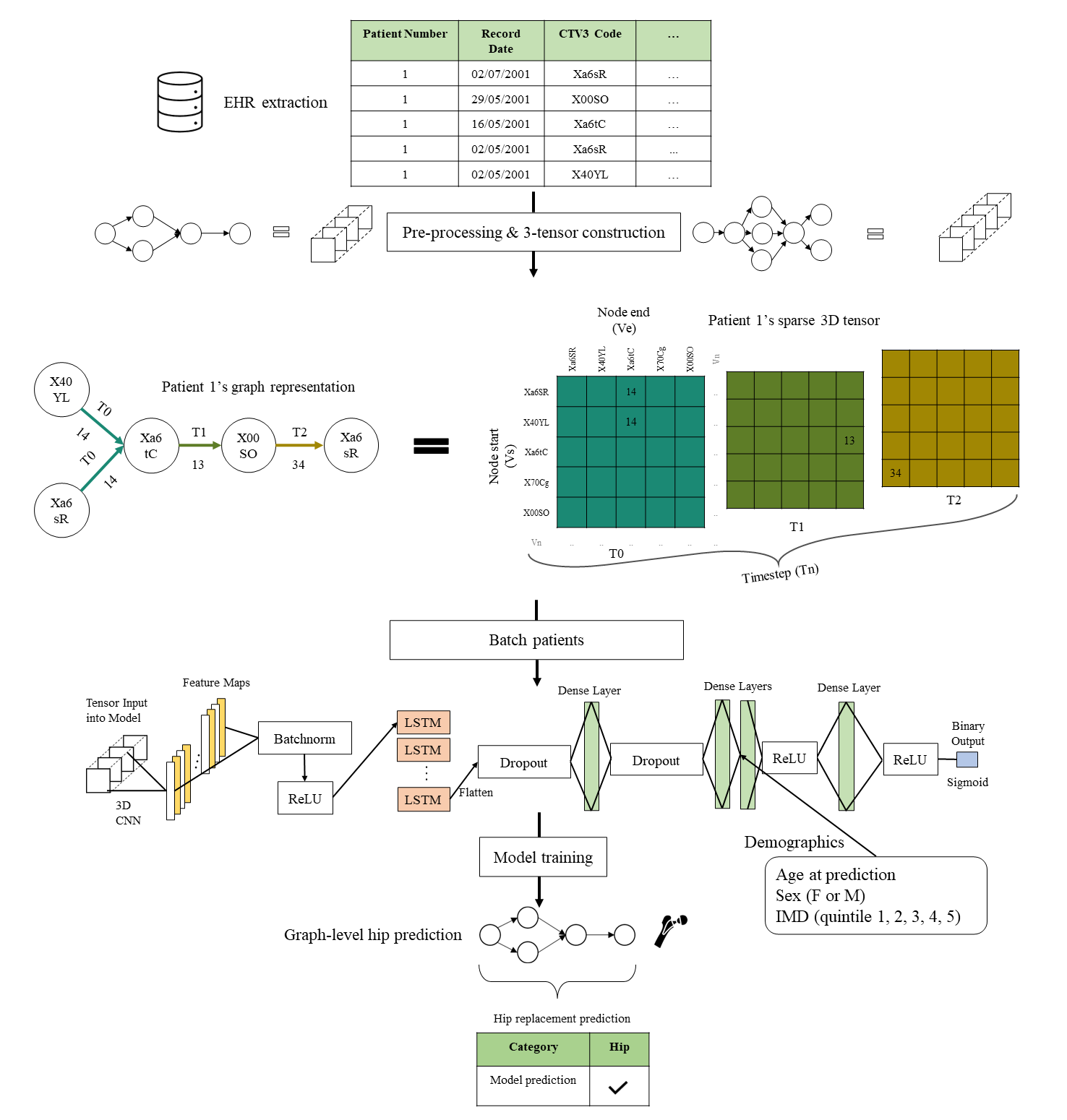}
    \caption{From raw data to hip replacement risk prediction. Here we show an example sequence for fictitious `Patient 1' of four Read Codes being recorded across three time steps (visits). Here there are only 5 Read Codes (nodes), however in reality the nodes span to over 500 codes.}
    \label{fig:data to prediction}
\end{figure*}

In discrete mathematics, a graph $G = (V,E)$ comprises nodes $V$ linked by edges $E$ \cite{Scarselli2009GraphNeuralNetworkModel,Hamilton2020GraphRepresentationLearning}. 
Our method involves transforming sequences of Read Codes from EHRs into temporal multigraphs, framing hip replacement risk prognosis as a graph classification problem. Each temporal graph represents an individual and is classified to provide risk probabilities.
We utilise Read Codes as graph nodes $V$, while temporal edges $E$ capture time intervals between code occurrences, measured in months, represented in a 3D tensor $G(i, j, k) = t_k$, where $i, j \in \{1,\dots, n\}$ indicate graph nodes, and $t_k$ is the time elapsed for the $k$th edge. Each of the 512 most frequently used CTV3 codes maps to one node $v_1,v_2,\dots,v_{512}$. 
Figure \ref{fig:data to prediction} demonstrates how we convert a snippet of an EHR into 3-tensors. We set $k = 100$ as the maximum number of time steps, resulting in a 3-tensor input size of $512 \times 512 \times 100$. 
By representing Read Codes as graphs rather than linear sequences, we accommodate multiple code recordings per patient visit, forming temporal graphs. These graphs are then fed into a \texttt{TG-CNN} deep learning algorithm, enabling pattern detection within Read Code graph structures and temporal features. An overview of using EHR data for joint replacement prediction is provided in Figure \ref{fig:data to prediction}, and the model architecture is detailed in Section \ref{sub:model architecture}.

\subsection{Model Architecture}
\label{sub:model architecture}

TensorFlow was employed to craft a custom 3D CNN Keras layer using sparse linear algebra. The temporal graph representation, sized at $512\times512\times100$, undergoes processing through this layer to extract Read Code sequence and elapsed time patterns (see Figure \ref{fig:data to prediction}).
The output of the CNN is a sequence of feature vectors capturing short patterns (accounting for elapsed time).
The CNN output is flattened, followed by batch normalisation for faster convergence, and subsequent layers include a Leaky Rectified Linear Unit (ReLU), LSTM (captures longer temporal patterns), dropout, dense layers, and concatenation with demographic features (if included). Finally, the model employs categorical cross-entropy loss with sigmoid for the categorical target.
Tensorflow 2.8.0, NumPy 1.19.2, Pandas 1.2.4, Scikit-Learn 0.23.1, and CUDA 10.2.89 were used on a desktop with a NVIDIA RTX 3090. Interested readers can access our Git repository\footnotemark[1] for code exploration. 

Events occurring further in the past may have a diminishing impact on recent events. Therefore, clinicians typically focus on recent health records rather than trawling through years' worth of data. Similarly, models predicting patient outcomes prioritise recent events. In this task, we limit our analysis to the 100 most recent primary care visits per patient to manage computational load. We front-pad any 3-tensors representing fewer than 100 visits, ensuring the most recent visits are always at the end of the 3-tensor.
Multi-stream networks are also explored to search over different granularities in time, in our two stream model we use a stride of 1 in the coarse stream and 2 in the fine stream.

Events occurring closely together may indicate a worse health condition compared to those with wider time intervals between them. To capture this, a modified version of elapsed time is stored, measuring the proximity of two events occurring. The graph is transformed using $G(i, j, k) = \exp(-\gamma t)$, where $\gamma$ is a trainable model parameter. This transformation assigns a score of 1 to simultaneous events and close to 0 for events with a large time gap. Elements indicating no interaction between node pairs at a set time are set to 0.

We employ an $\ell_1$ (LASSO) regularisation penalty on each filter to prevent overfitting and promote sparsity. This regularisation is applied by adding the $\ell_1$ norm to the loss: $loss + \lambda \sum_{i=0}^{n}|w_i|$, where $n$ is the number of features in the dataset, $w_i$ represents the weight for the $i$th feature, and $\lambda$ is the regularisation strength. Filters help approximate a subgraph, implying that if an individual's EHR is close to a given filter, their disease outcome will match. We incorporate a graph regularisation function, denoted as $\ell_G$, which penalises the model with a graph regularisation strength $\lambda_G$ if the filter deviates from a typical graph structure, such as when a node lacks incoming or outgoing connections to other nodes.

\subsection{Comparison Models}

An ablation study is carried out using 10 variations of the models to assess component benefit as shown in Table \ref{tab:tgcnn_ablation}, we also evaluate a model variation which includes prescriptions.

\begin{table}[h]
\caption{Components included in the ablation study models. Demo = demographics; Exp = exponential scaling; 2 str = two streams; $\ell_G$ = graph loss; $\ell_2$ = L2 regularization; $\ell_1$ = L1 regularization; LSTM = long short-term memory; Time = elapsed time; CNN = convolutional neural network.}
\label{tab:tgcnn_ablation}
\centering
\begin{tabular}{p{1.25cm}p{0.25cm}p{0.3cm}p{0.35cm}p{0.5cm}p{0.3cm}p{0.3cm}p{0.3cm}p{0.4cm}p{0.3cm}}
\toprule
\textbf{TGCNN} & \multicolumn{9}{c}{\textbf{Included}}\\ 
\midrule
Component & $\gamma$ & Exp & Demo & 2 str & $\ell_G$  & $\ell_2$ & $\ell_1$ & LSTM & Time \\
\midrule
Full & \checkmark & \checkmark & \checkmark & \checkmark & \checkmark & \checkmark & \checkmark & \checkmark & \checkmark \\
w/o $\gamma$ & & \checkmark & \checkmark & \checkmark & \checkmark & \checkmark & \checkmark & \checkmark & \checkmark \\
w/o Exp & & & \checkmark & \checkmark & \checkmark & \checkmark & \checkmark & \checkmark & \checkmark \\
w/o Time & \checkmark & \checkmark & \checkmark & \checkmark & \checkmark & \checkmark & \checkmark & \checkmark &  \\
w/o Demo & \checkmark & \checkmark &  & \checkmark & \checkmark & \checkmark & \checkmark & \checkmark & \checkmark \\
w/o 2 str & \checkmark & \checkmark & \checkmark &  & \checkmark & \checkmark & \checkmark & \checkmark & \checkmark \\
w/o $\ell_G$ & \checkmark & \checkmark & \checkmark & \checkmark &  & \checkmark & \checkmark & \checkmark & \checkmark \\
w/o $\ell_2$ & \checkmark & \checkmark & \checkmark & \checkmark & \checkmark &  & \checkmark & \checkmark & \checkmark \\
w/o $\ell_1$ & \checkmark & \checkmark & \checkmark & \checkmark & \checkmark & \checkmark &  & \checkmark & \checkmark \\
w/o LSTM & \checkmark & \checkmark & \checkmark & \checkmark & \checkmark & \checkmark & \checkmark &  & \checkmark \\
\bottomrule
\end{tabular}
\end{table}

When evaluating \texttt{TG-CNN} performance in modelling temporal dependencies within sequential graph data, robust baseline references are crucial. We compare RF and Logistic Regression (LR) with Recurrent Neural Network (RNN) and LSTM networks. RNNs specialise in processing sequential data by maintaining internal states over time, effectively capturing long-term dependencies within EHR data. However, their performance must be compared against simpler models like RF and LR, which lack explicit mechanisms for modelling such dependencies but offer simplicity and interpretability.

\subsection{Evaluation Approach} 
\label{sub:model evaluation}

We follow the Transparent Reporting of a Multivariate Prediction Model for Individual Prognosis or Diagnosis (TRIPOD)-AI statement for both the reporting of model development and the prediction models (see the Supplementary material for the completed checklist).

We took a random 10\% sample of the patients that met the inclusion criteria to test the trained model on, where this group represents the same proportion of the population that has a hip replacement as the full dataset. From the other 90\% we match every hip replacement patient to a control patient based on sex, IMD, and age at replacement. Note we do not oversample, each match is a new addition. Within the balanced training dataset, we perform 5-fold cross-validation when choosing hyperparameters (detailed in the supplied repository). We optimise the model based the on cross-validation mean validation accuracy value,
we conducted a random hyperparameter search on the Full \texttt{TG-CNN} model 20 times.
We take the model with the best AUROC score from the cross-validation and use the same hyperparameters across the ablation study models, running each model for a maximum of 12 hours. 
For the 10\% test dataset, we use half to recalibrate the model (we denote this as Test 1) and the other half (denoted as Test 2) to test the recalibrated model. 


Calibration is undertaken to assess how close the predicted probabilities reflect the true risk of needing a hip replacement. 
Calibration is vital to ensure that the predicted probabilities are accurate for different subgroups based on their risk levels \cite{Dhiman2022Method}.
We also assess discrimination (AUROC and Area under the Precision Recall Curve (AUPRC)) of the models.

After testing on the balanced data, we proceed to recalibrate the model. Since the model was trained on a balanced dataset, we adjust the model predictions to enhance alignment with the true incidence rates of the outcomes in the test set. For recalibration, we create a LR model that takes the linear predictors of the original model. The LR model improves calibration by estimating new coefficients. We can assess the calibration slope before and after recalibration to determine efficiency. 
Once the model is recalibrated on the Test 1 data, we validate the recalibrated model's performance on a second unseen test set (Test 2).
We resample the Test 2 data to assess model stability, bias, and variability using bootstrapping.
We also perform patient stratification on the best recalibrated \texttt{TG-CNN} model, assessing the model calibration on different demographic sub-groups separately. We create sub-groups based on sex (female and male), age at prediction (40-60, 60-70 and 70$+$ year olds) and IMD (quintiles).

\section{Results}

Table \ref{tab:hip_replace_cohort_summary} displays the characteristics of the cohort included in model training and testing. We conduct a Chi-squared analysis to compare the population of Test 2 to the National Joint Registry 12th Annual Report 2015, ensuring similar demographic distributions \cite{NJR2015}. Results in Table S2 in the supplementary materials show no significant difference in age, sex, and IMD between our Test 2 set and the NJR. However, there is a significant difference observed in weight category distribution between the datasets.

Table \ref{tab:imd_by_hip_replace} displays the distribution of hip replacements per deprivation quintile, indicating slightly higher rates in more deprived areas (IMD 1-3) compared to less deprived regions. No data are missing for age, sex, or IMD.

\begin{table}[h]
    \centering
    \caption{Hip replacement prediction model cohort characteristics. The BMI statistics are derived from the last recorded BMI measurement of each patient.}
    \begin{tabular}{lccc}
        & \textbf{Train}& \textbf{Test 1}& \textbf{Test 2}\\
        \textbf{Characteristic} & \textbf{($N$=18,374)}& \textbf{($N$=7,353)}&\textbf{($N$=7,353)}\\
         \midrule
         Sex: female &  11,626(63.3)&  4,422(60.2)&4,453(60.6)\\
         Hip replacements & 9,187(50) & 515(7.0) & 515(7.0) \\
         Age, y & & & \\
         \quad mean(std) &  72.5(8.5)&  67.9(9.3)&67.6(9.3)\\
         \quad median(min,max)&  73(53,89)&  67(53,89)&66(53,89)\\
         \quad mode & 76& 66& 66\\
         Age at replace, y & & &\\
         \quad mean(std) &  69.2(8.53)&  69.6(8.6)&69.4(8.8)\\
         \quad median(min,max)&  70(43,90)&  70(48,89)&70(48,88)\\
         \quad mode & 77& 78&77\\
         BMI & & &\\
         \quad Severely obese &  652(3.5)&  267(3.6)&285(3.9)\\
         \quad Obese &  5,803(31.6)&  2,035(27.7)&2,108(28.7)\\
         \quad Overweight &  6,721(36.6)& 2,694(36.6)&2,674(36.4)\\
         \quad Healthy weight &  4,304(23.4)&  1,716(23.3)&1,773(24.1)\\
         \quad Underweight &  287(1.6)&  102(1.4)&102(1.4)\\
         \quad Missing &  607(3.3)&  539(7.3)&411(5.6)\\
         IMD, quintile & & &
\\
         \quad 1 (most deprived) &  4,996(27.2)&  1,809(24.6)&1,565(21.3)\\
         \quad 2 &  4,640(25.3)&  1,665(22.6)&1,554(21.1)\\
         \quad 3 &  3,982(21.7)&  1,635(22.2)&1,373(18.7)\\
         \quad 4 &  2,648(14.4)&  1,223(16.6)&1,324(18.0)\\
         \quad 5 (least deprived) &  2,108(11.5)&  1,021(13.9)&1,537(20.9)\\
         Events Recorded  & & & \\
         \quad mean(std) & 10.6(13.5)& 24.7(22.8)& 26.7(23.5)\\
 \quad median(min,max)
& 6.0(2,161)& 18.0(2,224)& 20.0(2,182)\\
 \quad mode & 3& 8& 7\\
 \bottomrule
    \end{tabular}
    \label{tab:hip_replace_cohort_summary}
\end{table}

\begin{table}
    \centering
    \caption{Hip replacement occurrence in each IMD group N(\%).}
    \begin{tabular}{lccc}
        \toprule
         &  \textbf{Overall}&  \textbf{Test 1}& \textbf{Test 2}\\
         &  &   \textbf{(Uncalibrated)}&  \textbf{(Re-calibrated)}\\
         \midrule
         IMD 1 &  2,750(32.9)&  142(7.9)& 110(7.0)\\
         IMD 2 &  2,549(32.4)&  102(6.1)& 127(8.2)\\
         IMD 3 &  2,248(32.2)&  133(8.1)& 124(9.0)\\
         IMD 4 &  1,498(28.8)&  90(7.4)& 84(6.3)\\
         IMD 5 &  1,172(25.1)&  48(4.7)& 70(4.6)\\
         \bottomrule
    \end{tabular}
    \label{tab:imd_by_hip_replace}
\end{table}

\begin{figure}[h]
    \centering
    \includegraphics[width=0.73\linewidth]{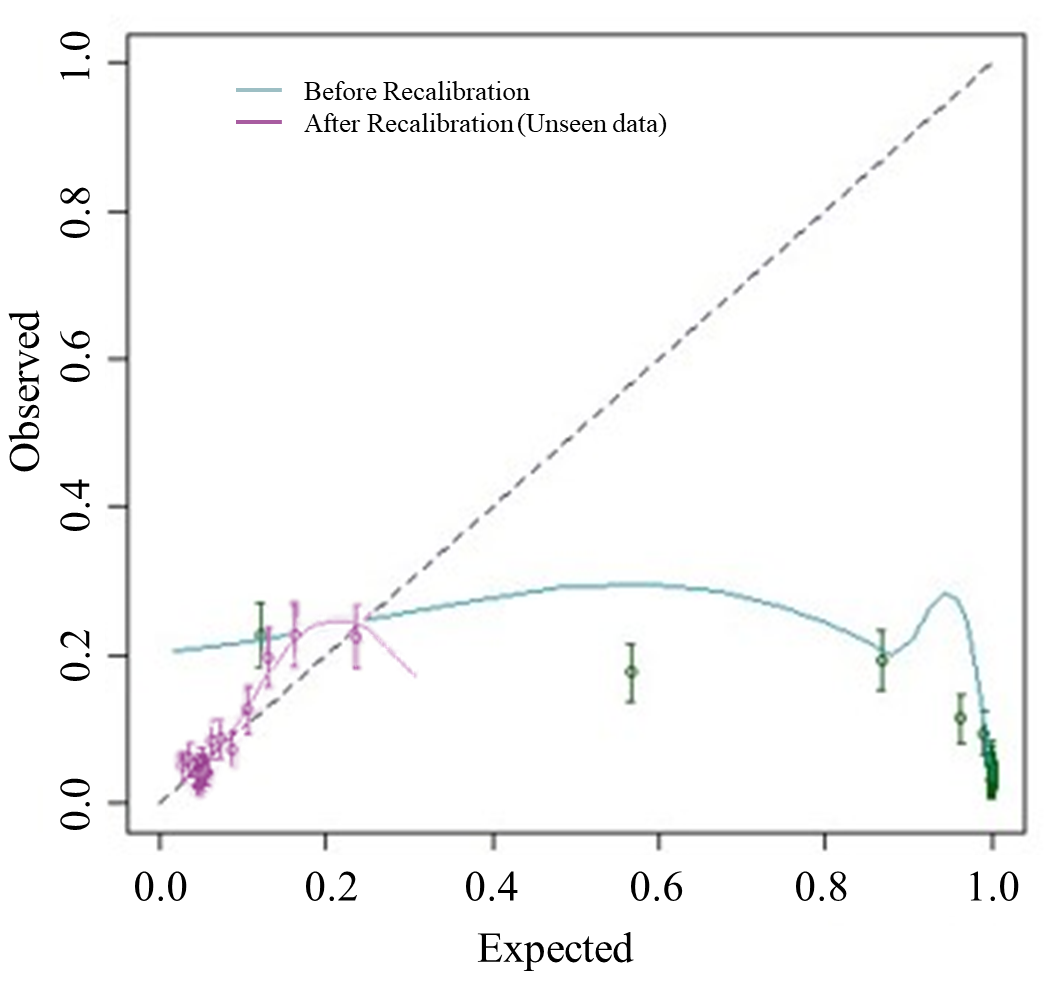}
    \caption{Calibration curve before and after recalibration (blue: before recalibration on Test 1 data, purple: after recalibration on the unseen (Test 2) data.}
    \label{fig:tgcnn_calcurve}
\end{figure}

Upon running the trained model on the test data, an inverted calibration curve was observed (Figure \ref{fig:tgcnn_calcurve}), with all probabilities below 0.5, indicating predictions skewed towards the positive group, despite around only 7\% of the population belonging to this group. However, recalibrating the model resulted in a closer fit to the optimal line and more accurate probability distribution. To understand this phenomenon, several evaluations were conducted. We tried training the model on an imbalanced dataset, where each case patient was matched with two control patients, yet the probability distribution remained skewed towards case patients. Further testing involved reserving 20\% of unmatched control patients for testing, resulting in similar probability skewing. This phenomenon persisted during training of baseline models, indicating it is data-based rather than model-based. Holding back a portion of matched training data as test data yielded good results even prior to calibration. These tests suggest that the occurrence is due to the model being well-trained on matched patients, resulting in different trajectories for unseen test patients.

Table \ref{tab:cv_auroc_and_cal_results} displays AUROC and calibration slope results from 5-fold cross-validation on the \texttt{TG-CNN} model. We select the w/o $\ell_1$ model with the validation C-slope value closest to 1 to illustrate before and after recalibration calibration curves (Figure \ref{fig:tgcnn_calcurve}), as this model demonstrates the best calibration.
The \texttt{TG-CNN} w/o exponential model outperformed others in terms of AUROC on the balanced dataset but showed poor recalibration, suggesting overfitting. Table \ref{tab:auroc_auprc_test2_ablation_results} and Table \ref{tab:auroc_auprc_test2_baseline_results} present AUROC and AUPRC results using the Test 2 (unseen) data on the \texttt{TG-CNN} and baseline models, respectively.

\begin{table}
    \centering
    \caption{AUROC and C-slope (mean (SD)) results for the models on the training set. Demo = demographics, Prescripts = prescriptions.}
    \begin{tabular}{p{1.9cm}p{1.25cm}p{1.25cm}p{1.25cm}p{1.25cm}}
    \toprule
     \textbf{TGCNN Model}&  \textbf{Train \hspace{1cm} AUROC} &  \textbf{Validation AUROC }&  \textbf{Train \hspace{1cm} C-slope}& \textbf{Validation C-slope}\\
    \midrule
     w/o exp&  0.97 (0.00)&  0.94 (0.01)
&  1.00 (0.01)& 0.82 (0.06)
\\
     w/o $\ell_1$&  0.94 (0.01)
&  0.88 (0.01)
&  1.04 (0.02)& 1.00 (0.07)\\
     w/o LSTM
&  0.92 (0.00)
&  0.86 (0.02)
&  0.84 (0.15)& 0.89 (0.07)
\\
     w/o $\ell_G$&  0.97 (0.01)&  0.86 (0.01)
&  1.04 (0.05)& 0.61 (0.06)
\\
     w/o demos
&  0.96 (0.04)
&  0.85 (0.01)
&  1.06 (0.02)& 0.68 (0.11)
\\
     Full
&  0.97 (0.02)&  0.85 (0.00)
&  1.02 (0.03)& 0.69 (0.13)
\\
     w/o elapsed time
& 0.97 (0.02)& 0.84 (0.01)
& 0.10 (0.05)&0.59 (0.08)
\\
     w/o $\ell_2$& 0.97 (0.01)
& 0.84 (0.02)& 1.02 (0.03)&0.60 (0.12)
\\
     w/o two streams& 0.52 (0.00)
& 0.52 (0.01)
& 0.87 (0.25)&0.86 (0.60)
\\
     w/o $\gamma$& 0.52 (0.00)
& 0.52 (0.01)
& 1.07 (0.15)&0.96 (0.56)
\\

 with prescripts& 0.52 (0.00)& 0.52 (0.01)& 0.82 (0.19)&0.68 (0.45)\\
 \bottomrule
    \end{tabular}
    \label{tab:cv_auroc_and_cal_results}
\end{table}

\begin{table}
    \centering
    \caption{AUROC and AUPRC results for the recalibrated \texttt{TG-CNN} models on Test 2 data.}
    \begin{tabular}{p{1.9cm}p{1.25cm}p{1.25cm}p{1.25cm}p{1.25cm}}
    \toprule
     \textbf{TGCNN Model}&  \textbf{AUROC mean (SD)}&  \textbf{AUROC (95\% CI)}&  \textbf{AUPRC mean (SD)}& \textbf{AUPRC (95\% CI)}\\
     \midrule
 w/o $\ell_2$ & 0.72 (0.01)& (0.72,~0.73)& 0.19 (0.01)&(0.16,~0.21)\\
 w/o LSTM& 0.72 (0.01)& (0.71,~0.72)& 0.16 (0.01)&(0.14,~0.18)\\
 w/o $\ell_1$ & 0.72 (0.02)& (0.71,~0.73)& 0.22 (0.03)&(0.16,~0.28)\\
 w/o $\ell_G$& 0.70 (0.02)& (0.69,~0.71)& 0.17 (0.01)&(0.15,~0.18)\\
 w/o demos& 0.70 (0.02)& (0.68,~0.71)& 0.15 (0.02)&(0.12,~0.19)\\
 Full& 0.69 (0.02)& (0.68,~0.70)& 0.17 (0.01)&(0.14,~0.19)\\
 with prescripts&  0.68 (0.01)&  (0.68,~0.69)
&  0.12 (0.00)& (0.11,~0.13)\\
     w/o elapsed time&  0.68 (0.01)
&  (0.67,~0.69)&  0.15 (0.01)& (0.12,~0.18)\\
 w/o exp& 0.63 (0.02)& (0.62,~0.65)& 0.17 (0.01)&(0.15,~0.19)\\
     w/o $\gamma$&  0.59 (0.01)&  (0.58,~0.60)
&  0.08 (0.01)& (0.07,~0.09)
\\
     w/o two streams&  0.57 (0.01)&  (0.56,~0.58)&  0.08 (0.01)& (0.07,~0.09)\\
    \bottomrule
    \end{tabular}
    \label{tab:auroc_auprc_test2_ablation_results}
\end{table}

\begin{table}[h]
    \centering
    \caption{Results for the recalibrated baseline models on the unseen test set.}
    \begin{tabular}{p{2.15cm}p{1.15cm}p{1.15cm}p{1.15cm}p{1.15cm}}
    \toprule
     \textbf{Model}&  \textbf{AUROC mean(std)}&  \textbf{AUROC (95\% CI)}&  \textbf{AUPRC mean(std)}&\textbf{AUPRC (95\% CI)}\\
    \midrule
    RNN &  0.70~(0.01)&  (0.69,~0.71)&  0.14~(0.01)& (0.12,~0.16)\\
    LSTM &  0.65~(0.01)&  (0.64,~0.66) &  0.12~(0.01)& (0.10,~0.14)\\
    LR demo only&  0.58~(0.02)&  (0.57,~0.59)&  0.08~(0.00)& (0.07,~0.09)\\
     RF demo only&  0.58~(0.01)&  (0.57,~0.58)&  0.08~(0.00) & (0.07,~0.09)\\
    LR CTV3 only&  0.53~(0.02)&  (0.52,~0.54)&  0.09~(0.01)& (0.07,~0.11)\\
     LR demo and CTV3&  0.53~(0.01)&  (0.52,~0.54)&  0.09~(0.01)& (0.08,~0.10)\\
     RF CTV3 only&  0.53~(0.01)&  (0.52,~0.54)&  0.09~(0.00)& (0.08,~0.09)\\
     RF demo and CTV3&  0.52~(0.01)&  (0.51,~0.53)&  0.08~(0.01)& (0.07,~0.10)\\
     
 \bottomrule
    \end{tabular}
    \label{tab:auroc_auprc_test2_baseline_results}
\end{table}


The most important features, according to the best RF model are provided in the supplementary materials Figure S1. Additionally, Figure S2 in the supplementary materials displays the top 10 largest odds ratios from the LR model, indicating variables that act as risk multipliers. As anticipated, both models seem to recognise that Read Codes associated to hip pain and OA are crucial for predicting hip replacement.

We conduct subgroup analysis (patient stratification) on the recalibrated \texttt{TG-CNN} w/o $\ell_1$ model. See the supplementary materials for calibration curves (Figures S3 - S12) which demonstrate fair equity across demographic subgroups. There seems to be slightly more underprediction in the female population compared to males, but overall, predicted risks are reasonable in a large proportion of the population. Younger populations tend to experience more overprediction, while individuals aged 60-70 exhibit the best calibration; however, those over 70 are notably underpredicted. Regarding deprivation levels, IMD 1, 2, and 4 demonstrate the best calibration, while IMD 5 overpredicts hip replacement and IMD underpredicts, indicating no bias based on IMD.

\section{Discussion} 

The RF and LR models, though interpretable with top predictors like hip pain, hip OA, OA, and tobacco smoking, showed poor AUROC and AUPRC. The RF model emphasised OA, smoking, and weight. Using demographics alone improved performance slightly compared to using Read Codes in the recalibrated RF and LR models. While these models performed competitively on the balanced training set, their recalibrated application to Test 2 data led to poorer AUPRC results.
Excluding demographics in the \texttt{TG-CNN} model resulted in overfitting, indicating that including predictors beyond basic demographic information may be beneficial. 

The LSTM and RNN baseline models might perform poorly due to time decay on events occurring further in the past, weakening relationships with longer time intervals. Reversing the input vector prevented convergence, indicating that the model may learn best from a mix of recent and historical events. With simple RNNs, we cannot include elapsed time, so we only feed the events into the model as a sequence, losing some temporal features. Including an LSTM in the \texttt{TG-CNN} model could be beneficial for retaining distant Read Codes in memory rather than focusing solely on recent ones.

Using an exponential function for elapsed time allows rescaling to avoid extreme values in neural networks and mitigate potential under/overflow issues with half-precision arithmetic. The negative exponential ensures that actions taken quickly in succession have values close to 1, while those with greater temporal gaps are closer to 0. 
The model without the exponential component achieved the highest validation AUROC with minimal difference from the training set, suggesting no overfitting occurred. However, despite its success on balanced data during training, it performed poorly during recalibration, leading to significant underprediction for some patients. This suggests that the exponential component improves generalisability to unbalanced datasets. It also implies that the time step (axis $z$ of the 3D tensor representing the temporal graph) requires the exponential component to recognise past events compared to recent ones.

Incorporating prescriptions tripled the number of eligible records for our model. Including prescriptions in the model may result in poor validation AUROC outcomes because the graphs become inundated with prescriptions rather than diagnostic Read Codes. 
Moreover, additional features could increase model complexity, potentially prolonging convergence time and yielding poorer results.

Elastic Net might not perform as effectively as $\ell_2$ regularisation alone, especially when handling highly correlated features, as indicated by the good performance of our \texttt{TG-CNN} model without $\ell_1$ regularisation. $\ell_1$ regularisation tends to select only one feature from a group of correlated features, effectively disregarding the rest. In contrast, $\ell_2$ regularisation penalises all feature coefficients, distributing the weight more evenly among correlated features.
Our custom graph regularisation function $\ell_G$ seems to effectively prevent overfitting.

The model without a second stream would not converge, this shows that having a long and a short stream is useful for seeing both large and small features.
Having two different kernel sizes (in this paper we refer to them as `streams') capture features at different scales. A fine branch with smaller kernel sizes might capture finer details and textures, while a coarse branch with larger kernel sizes captures broader patterns and structures. Combining these branches allows the model to learn features at multiple scales simultaneously, enhancing its ability to understand complex patterns in the data. 

The \texttt{TG-CNN} model without elapsed time overfitted and exhibited a poor C-slope value, highlighting the significance of incorporating irregular elapsed time between Read Code recordings to enhance predictive performance.

The \texttt{TG-CNN} model utilises EHR data from primary care, allowing for the identification of individuals at higher risk of future hip replacement. With this model, targeted preventative care could be applied to slow progression, severity, or pain. Future research should investigate clinical outcomes and potential interventions in primary care based on this model.

The \texttt{TG-CNN} model serves as an alternative to the current EHR-based hip prediction model \cite{Yu2019HipKneeReplace}, enabling the infusion of temporality and trajectory. In future work, we aim to incorporate explainability into this approach using the filters from the CNN layer, to enable further trust in the model and 
explain why certain predictions may have occurred \cite{Pulikottil2020}. We also aim to reconfigure this model to determine the risk of needing a hip replacement 5-years in advance, allowing clinicians more time to apply interventions and planning.

Limitations arise from GPs potentially lacking time to code diagnoses accurately during consultations, leading to misclassification. 
Although multi-modal models, such as those incorporating imaging, could be beneficial, time and cost constraints in healthcare favour models quicker to train, especially when recalibration is necessary to prevent temporal drift. Utilising readily available EHRs data facilitates implementation in routine primary care.

The absence of limb sidedness/laterality data prevents specifying which leg has undergone replacement, potentially resulting in misinterpretation, although using the first record of primary replacement attempts to mitigate this. The primary care data used in the study may lack the depth found in secondary care records, which could provide more comprehensive replacement details and reasons for the replacements, such as trauma or elective surgeries. Left censoring issues, stemming from having Read Codes only from a specific date onwards, might overlook joint replacements conducted before the analysis period. 

\section{Conclusion}

The \texttt{TG-CNN} model can aid in clinical decision-making by targeting individuals with an elevated risk of future hip replacement for intensive non-surgical management or active monitoring, enabling the application of preventive treatments and care. Implementing this model clinically may potentially reduce patients' time in pain, enhance quality of life, and improve healthcare efficiency and resource allocation.

\section*{Ethical Approval}

The study was approved by the University of Leeds School of Medicine Research Ethics Committee (SoMREC/13/079) and the ResearchOne Project Committee (201428378A).

\section*{Acknowledgement}

ZH is supported through funding by the EPSRC (Grant No. \href{https://gtr.ukri.org/projects?ref=EP%2FS024336%2F1}{EP/S024336/1}). PGC and SRK are funded in part by the National Institute for Health and Care Research (NIHR) through the Leeds Biomedical Research Centre.



\bibliography{references}

\begin{thebibliography}{10}
\providecommand{\url}[1]{#1}
\csname url@samestyle\endcsname
\providecommand{\newblock}{\relax}
\providecommand{\bibinfo}[2]{#2}
\providecommand{\BIBentrySTDinterwordspacing}{\spaceskip=0pt\relax}
\providecommand{\BIBentryALTinterwordstretchfactor}{4}
\providecommand{\BIBentryALTinterwordspacing}{\spaceskip=\fontdimen2\font plus
\BIBentryALTinterwordstretchfactor\fontdimen3\font minus \fontdimen4\font\relax}
\providecommand{\BIBforeignlanguage}[2]{{%
\expandafter\ifx\csname l@#1\endcsname\relax
\typeout{** WARNING: IEEEtran.bst: No hyphenation pattern has been}%
\typeout{** loaded for the language `#1'. Using the pattern for}%
\typeout{** the default language instead.}%
\else
\language=\csname l@#1\endcsname
\fi
#2}}
\providecommand{\BIBdecl}{\relax}
\BIBdecl

\bibitem{WhittyCMOReport2020}
\BIBentryALTinterwordspacing
C.~Whitty, ``{Chief Medical Officer’s annual report 2020: health trends and variation in England},'' NHS, Tech. Rep., 2020. [Online]. Available: \url{https://www.gov.uk/government/publications/chief-medical-officers-annual-report-2020-health-trends-and-variation-in-england}
\BIBentrySTDinterwordspacing

\bibitem{chen2023temporal}
X.~Chen, H.~Tang, J.~Lin, and R.~Zeng, ``Temporal trends in the disease burden of osteoarthritis from 1990 to 2019, and projections until 2030,'' \emph{PLoS One}, vol.~18, no.~7, p. e0288561, 2023.

\bibitem{Conaghan2014}
P.~G. Conaghan, M.~Kloppenburg, G.~Schett, and J.~W.~J. Bijlsma, ``{Osteoarthritis research priorities : a report from a EULAR ad hoc expert committee},'' pp. 1442--1445, 2014.

\bibitem{Arslan2022KneeOAIncidence}
I.~G. Arslan, J.~Damen, M.~Wilde, J.~J. Driest, P.~J. Bindels, J.~Lei, D.~Schiphof, and S.~M. Bierma‐Zeinstra, ``{Incidence and prevalence of knee osteoarthritis using codified and narrative data from electronic health records: a population‐based study},'' \emph{Arthritis Care {\&} Research}, pp. 0--3, 2022.

\bibitem{Cook2023FrailtyDeprivation}
M.~J. Cook, M.~Lunt, D.~M. Ashcroft, T.~Board, and T.~W. O’Neill, ``{The Impact of Frailty and Deprivation on the Likelihood of Receiving Primary Total Hip and Knee Arthroplasty among People with Hip and Knee Osteoarthritis},'' \emph{Journal of Frailty and Aging}, vol.~12, no.~4, pp. 298--304, 2023.

\bibitem{Keavy2022MSKPrev}
R.~Keavy, ``{The prevalence of musculoskeletal presentations in general practice: an epidemiological study},'' \emph{The British journal of general practice : the journal of the Royal College of General Practitioners}, vol.~70, pp. 1--7, 2020.

\bibitem{VersusArthritis2023}
\BIBentryALTinterwordspacing
{Versus Arthritis}, ``{The State of Musculoskeletal Health 2023 Arthritis and Other Musculoskeletal Conditions in Numbers},'' Versus Arthritis, Tech. Rep., 2023. [Online]. Available: \url{https://www.versusarthritis.org/media/25650/versus-arthritis-state-msk-musculoskeletal-health-2023-accessible.docx#:~:text=Every year 1 IN 5,to help manage their symptoms.}
\BIBentrySTDinterwordspacing

\bibitem{NJR2015}
\BIBentryALTinterwordspacing
{The NJR Editorial Board}, ``{National Joint Registry 12th Annual Report},'' National Joint Registry, Tech. Rep., 2015. [Online]. Available: \url{https://reports.njrcentre.org.uk/Portals/3/PDFdownloads/NJR 12th Annual Report 2015.pdf}
\BIBentrySTDinterwordspacing

\bibitem{Hobbs2016ClinicalWorkload}
F.~D. Hobbs, C.~Bankhead, T.~Mukhtar, S.~Stevens, R.~Perera-Salazar, T.~Holt, and C.~Salisbury, ``{Clinical workload in UK primary care: a retrospective analysis of 100 million consultations in England, 2007–14},'' \emph{The Lancet}, vol. 387, no. 10035, pp. 2323--2330, 2016.

\bibitem{Zhang2019ATTAIN}
Y.~Zhang, X.~Yang, J.~Ivy, and M.~Chi, ``{Attain: Attention-based time-aware LSTM networks for disease progression modeling},'' \emph{IJCAI International Joint Conference on Artificial Intelligence}, vol. 2019-Augus, pp. 4369--4375, 2019.

\bibitem{Wu2022OA-MedSQL}
T.~Wu, Y.~Wang, Y.~Wang, E.~Zhao, and G.~Wang, ``{OA-MedSQL: Order-Aware Medical Sequence Learning for Clinical Outcome Prediction},'' in \emph{2021 IEEE International Conference on Bioinformatics and Biomedicine (BIBM)}.\hskip 1em plus 0.5em minus 0.4em\relax IEEE, 2022, pp. 1585--1589.

\bibitem{Chiew2020}
C.~J. Chiew, N.~Liu, T.~H. Wong, Y.~E. Sim, and H.~R. Abdullah, ``{Utilizing Machine Learning Methods for Preoperative Prediction of Postsurgical Mortality and Intensive Care Unit Admission},'' vol. 272, no.~6, pp. 1133--1139, 2020.

\bibitem{Hancox2022}
Z.~Hancox and S.~D. Relton, ``{Temporal Graph-Based CNNs (TG-CNNs) for Online Course Dropout Prediction},'' \emph{Lecture Notes in Computer Science (including subseries Lecture Notes in Artificial Intelligence and Lecture Notes in Bioinformatics)}, vol. 13515 LNAI, no.~Ml, pp. 357--367, 2022.

\bibitem{Yu2019HipKneeReplace}
D.~Yu, K.~P. Jordan, K.~I. Snell, R.~D. Riley, J.~Bedson, J.~J. Edwards, C.~D. Mallen, V.~Tan, V.~Ukachukwu, D.~Prieto-Alhambra, C.~Walker, and G.~Peat, ``{Development and validation of prediction models to estimate risk of primary total hip and knee replacements using data from the UK: Two prospective open cohorts using the UK Clinical Practice Research Datalink},'' \emph{Annals of the Rheumatic Diseases}, vol.~78, no.~1, pp. 91--99, 2019.

\bibitem{Leung2020}
K.~Leung, B.~Zhang, J.~Tan, Y.~Shen, K.~J. Geras, J.~S. Babb, K.~Cho, G.~Chang, and C.~M. Deniz, ``{Prediction of total knee replacement and diagnosis of osteoarthritis by using deep learning on knee radiographs: Data from the osteoarthritis initiative},'' \emph{Radiology}, vol. 296, no.~3, pp. 584--593, 2020.

\bibitem{Yi2024}
Y.~Xu, X.~Hao, L.~Weixuan, H.~Liuu, J.~Guo, W.~Wang, H.~Ruan, Z.~Sun, and C.~Fan, ``{Development and Validation of a Deep-Learning Model to Predict Total Hip Replacement on Radiographs},'' \emph{The Journal of Bone and Joint Surgery}, vol. 106, no.~5, pp. 389--396, 2024.

\bibitem{Sun2020IrregularTimeReview}
\BIBentryALTinterwordspacing
C.~Sun, S.~Hong, M.~Song, and H.~Li, ``{A Review of Deep Learning Methods for Irregularly Sampled Medical Time Series Data},'' pp. 1--20, 2020. [Online]. Available: \url{http://arxiv.org/abs/2010.12493}
\BIBentrySTDinterwordspacing

\bibitem{Smith2020UnderstandingEngland}
C.~Smith, J.~Hewison, R.~M. West, S.~R. Kingsbury, and P.~G. Conaghan, ``{Understanding patterns of care for musculoskeletal patients using routinely collected National Health Service data from general practices in England},'' \emph{Health Informatics Journal}, vol.~26, no.~4, pp. 2470--2484, 12 2020.

\bibitem{IMD2019}
\BIBentryALTinterwordspacing
M.~o. H.~C. Government and Local, ``{The English Indices of Deprivation 2019 - Technical Report},'' Tech. Rep., 2019. [Online]. Available: \url{https://www.gov.uk/government/publications/english-indices-of-deprivation-2019- technical-report}
\BIBentrySTDinterwordspacing

\bibitem{Scarselli2009GraphNeuralNetworkModel}
F.~Scarselli, M.~Gori, A.~C. Tsoi, M.~Hagenbuchner, and G.~Monfardini, ``{The graph neural network model},'' \emph{IEEE Transactions on Neural Networks}, vol.~20, no.~1, pp. 61--80, 2009.

\bibitem{Hamilton2020GraphRepresentationLearning}
W.~L. Hamilton, ``{Graph Representation Learning},'' \emph{Synthesis Lectures on Artificial Intelligence and Machine Learning}, vol.~14, no.~3, pp. 1--159, 9 2020.

\bibitem{Dhiman2022Method}
P.~Dhiman, J.~Ma, C.~L. Andaur~Navarro, B.~Speich, G.~Bullock, J.~A. Damen, L.~Hooft, S.~Kirtley, R.~D. Riley, B.~Van~Calster, K.~G. Moons, and G.~S. Collins, ``{Methodological conduct of prognostic prediction models developed using machine learning in oncology: a systematic review},'' \emph{BMC Medical Research Methodology}, vol.~22, no.~1, pp. 1--16, 2022.

\bibitem{Pulikottil2020}
S.~C. Pulikottil and M.~Gupta, ``{ONet - A Temporal Meta Embedding Network for MOOC Dropout Prediction},'' \emph{Proceedings - 2020 IEEE International Conference on Big Data, Big Data 2020}, pp. 5209--5217, 2020.

\end{thebibliography}

\end{document}